\pdfoutput=1

\documentclass[11pt]{article}

\usepackage[final]{acl}

\usepackage{times}
\usepackage{latexsym}

\usepackage[T1]{fontenc}

\usepackage[utf8]{inputenc}

\usepackage{microtype}

\usepackage{inconsolata}

\usepackage{algorithm}
\usepackage{algorithmic}
\usepackage{amsmath}
\usepackage{bbm}
\usepackage{multicol}
\usepackage{hyperref}
\usepackage[nameinlink]{cleveref}

\usepackage{amssymb}
\usepackage{pifont}

\usepackage{color}
\usepackage{tabularray}

\usepackage{newunicodechar}

\usepackage{caption}
\usepackage{subcaption}

\usepackage{blindtext}
\usepackage{graphicx}
\usepackage{makecell}

\usepackage{booktabs}
\usepackage[normalem]{ulem}
\useunder{\uline}{\ul}{}

\usepackage{lipsum}
\usepackage{tikz}
\usepackage{fancyvrb}
\usepackage{listings}
\usepackage{enumitem}
\usepackage{tcolorbox}
\usepackage{multicol}
\usepackage{siunitx}
\usepackage{xpatch}
\usepackage{arydshln} 

\usepackage{multirow}  
\usepackage{pifont}
\usepackage{amsmath}

\DeclareMathOperator*{\argmax}{arg\,max}

\newcommand\model{TestNUC}

\title{TestNUC: Enhancing Test-Time Computing Approaches and Scaling through Neighboring Unlabeled Data Consistency}

\author{
 \textbf{Henry Peng Zou\textsuperscript{1}\thanks{Equal Contribution.}},
 \textbf{Zhengyao Gu\textsuperscript{1}\footnotemark[1]},
 \textbf{Yue Zhou\textsuperscript{1}},
 \textbf{Yankai Chen\textsuperscript{2}},
 \textbf{Weizhi Zhang\textsuperscript{1}},
\\
 \textbf{Liancheng Fang\textsuperscript{1}},
 \textbf{Yibo Wang\textsuperscript{1}},
 \textbf{Yangning Li\textsuperscript{3}},
 \textbf{Kay Liu\textsuperscript{1}},
 \textbf{Philip S. Yu\textsuperscript{1}}
\\
 \textsuperscript{1}University of Illinois Chicago,
 \textsuperscript{2}Cornell University
 \textsuperscript{3}Tsinghua University
\\
 \texttt{
  \{pzou3, zgu24\}@uic.edu
 }
}

\begin{document}
\maketitle
\begin{abstract}
Test-time computing approaches, which leverage additional computational resources during inference, have been proven effective in enhancing large language model performance. This work introduces a novel, linearly scaling approach, TestNUC, that improves test-time predictions by leveraging the local consistency of neighboring unlabeled data-it classifies an input instance by considering not only the model's prediction on that instance but also on neighboring unlabeled instances. We evaluate TestNUC across eight diverse datasets, spanning intent classification, topic mining, domain discovery, and emotion detection, demonstrating its consistent superiority over baseline methods such as standard prompting and self-consistency. Furthermore, TestNUC can be seamlessly integrated with existing test-time computing approaches, substantially boosting their performance. Our analysis reveals that TestNUC scales effectively with increasing amounts of unlabeled data and performs robustly across different embedding models, making it practical for real-world applications. Our code is available at \textcolor{blue}{\url{https://github.com/HenryPengZou/TestNUC}}.
\end{abstract}

\begin{figure}[!t]
    \centering
    \includegraphics[width=\columnwidth]{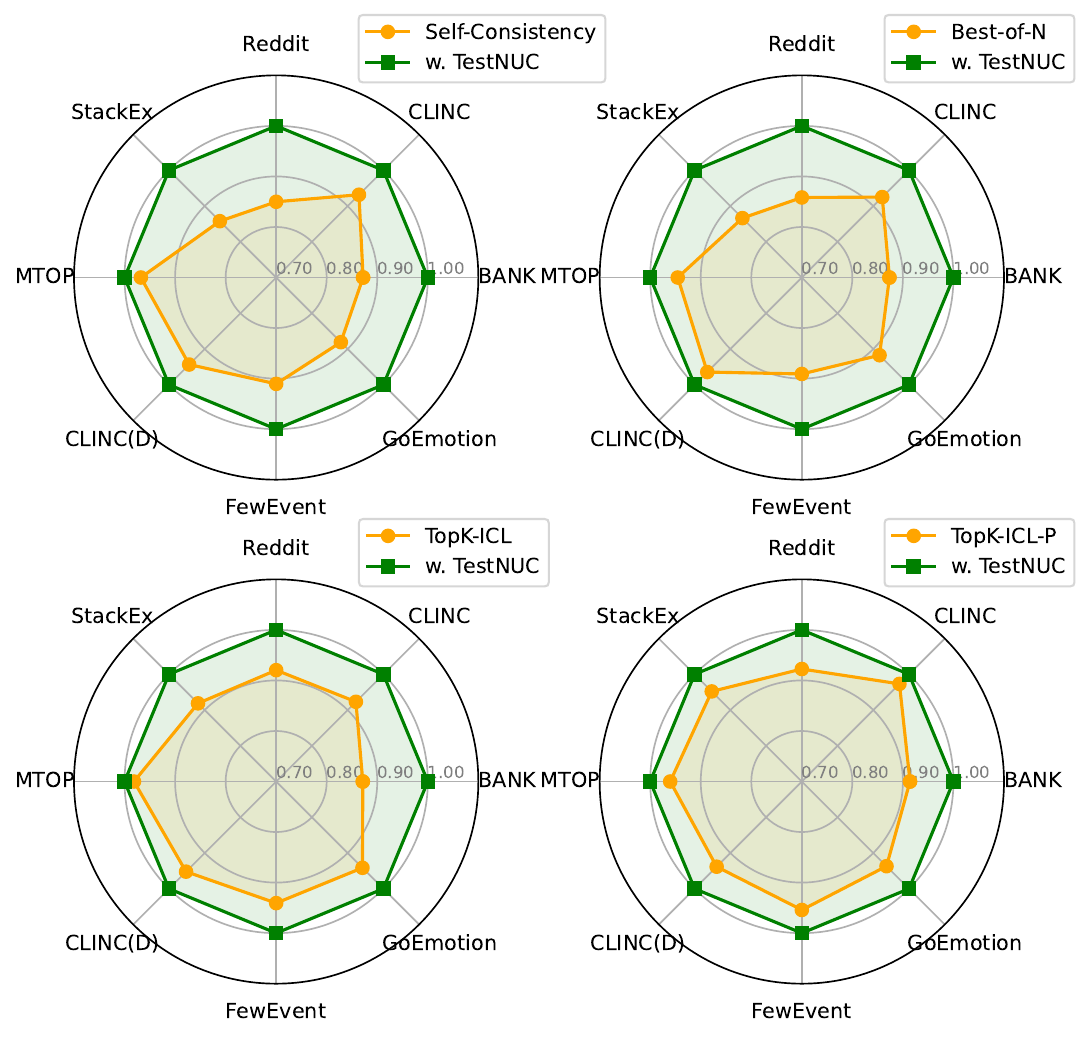} 
    \caption{TestNUC effectively integrates with both \textit{output-level} (e.g., Self-Consistency, Best-of-N) and \textit{input-level} (e.g., ICL-based) test-time computing methods, consistently enhancing their performance across eight datasets. More details in Section \ref{sec:experiment} and Table \ref{tab:main_enhance}.}
    \label{fig:radar_plot}
\end{figure}

\section{Introduction}

Test-time computing approaches, which leverage additional computational resources during inference to 
enhance performance, have gained increasing attention in the era of large language models (LLMs) \cite{test-time-compute, dong2024survey}.
There are two primary strategies for modifying an LLM’s distribution at test time: (1) \textbf{at the \textit{input level}}: augmenting the prompt with additional tokens (e.g., few-shot in-context learning \cite{mosbach-etal-2023-shot}); or (2) \textbf{at the \textit{output level}}: sampling multiple candidate answers and aggregating them (e.g., self-consistency \cite{self-consistency}, best-of-N \cite{beeching2024scalingtesttimecompute}).
Despite demonstrating promising capabilities, input-augmentation approaches incur a computational cost that scales \textit{quadratically} with the number of added tokens in the prompt, making them more computationally expensive than output-sampling methods. Meanwhile, output-sampling approaches typically overlook the potential of \textit{large amounts of unlabeled data} that are often available in real-world settings \cite{berthelot2019mixmatch, sohn2020fixmatch, zou-caragea-2023-jointmatch, zou2025gleangeneralizedcategorydiscovery, gu2025semi}.

\begin{figure}[!t]
    \centering
    \includegraphics[width=\columnwidth]{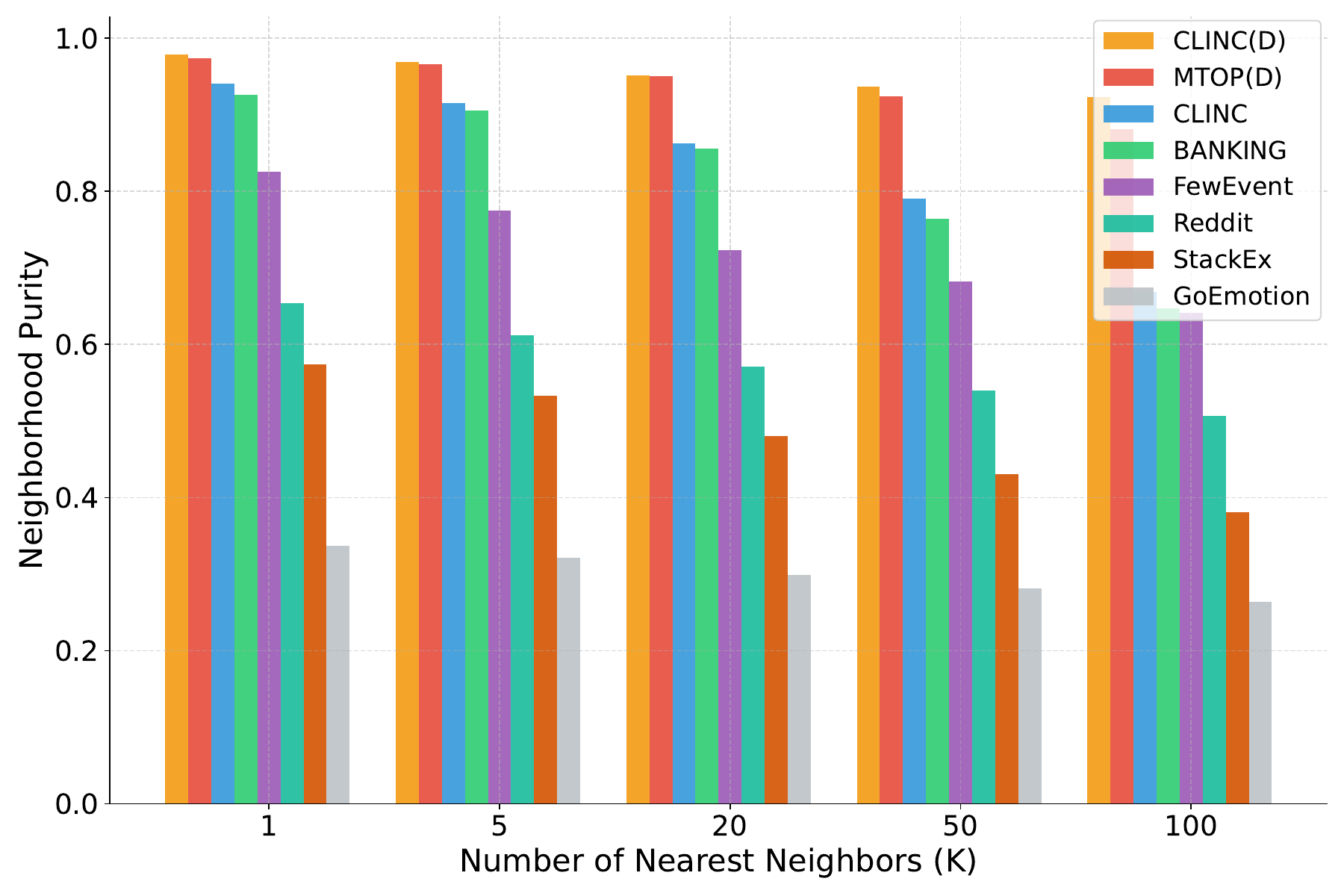} 
    \caption{Neighboring samples tend to be instances of the same semantic class.}
    \label{fig:analysis_neighborhood_purity}
\end{figure}

To bridge these gaps, we present an initial exploration of how unlabeled data can be efficiently leveraged to enhance test-time computing approaches. We hypothesize that instances with similar embeddings are likely to share the same semantic label, which can provide unsupervised signals for improving inference consistency, particularly for challenging instances \cite{van2020scan}. Our pilot experiments across various benchmarks reveal strong semantic label consistency among neighboring instances, and we find that aggregating these neighborhood labels through simple aggregation methods such as majority voting leads to stable and accurate predictions (as shown in Figure \ref{fig:analysis_neighborhood_purity}, \ref{fig:analysis_majority_vote} in Section \ref{sec:analysis}).

Motivated by these findings, we propose \textbf{\model}, a simple yet effective approach that enhances test-time LLM predictions by leveraging neighboring unlabeled data consistency. Concretely, TestNUC consists of two key steps: \ding{182} \textbf{Neighbor Retrieval}, where we identify the top-K nearest unlabeled neighbors of a test sample based on feature similarity; and \ding{183} \textbf{Collaborative Prediction}, where the LLM generates predictions for both the test sample and its retrieved neighbors, which are then aggregated to obtain the final answer. 
The intuition behind \model~is that samples in close proximity within the embedding space are likely to share similar labels.
By incorporating predictions of nearby unlabeled samples, the LLM can exploit the consistency of local data structures to better contextualize and refine its decision-making, effectively using unlabeled examples as an auxiliary signal to boost test-time performance while reducing noise and uncertainty \cite{van2020scan, zhou-etal-2024-paraphrase, wang2025bangs}.

We evaluate our approach across diverse tasks, including intent classification, topic mining, domain discovery, and emotion detection, using eight datasets that cover a wide spectrum of granularities, with class sizes ranging from 10 to 150. 
Our results demonstrate that \model~consistently outperforms baseline methods, such as standard prompting and self-consistency \cite{self-consistency}, by a large margin across four large language models, showing its effectiveness in leveraging unlabeled data for test-time computation. 
Moreover, \model~can be seamlessly integrated with existing test-time computing approaches, such as TopK-ICL \cite{peng-etal-2024-revisiting, gao2024on}, best-of-N \cite{lightman2024lets, beeching2024scalingtesttimecompute} and self-consistency \cite{self-consistency}, significantly boosting their performances (as illustrated in Figure \ref{fig:radar_plot}). 
In addition, \model~is effective across various embedders of different sizes and scales well with increasing amounts of unlabeled data (as shown in Figure \ref{fig:influence_num_unlabeled_data_log}), making it applicable to real-world scenarios.

\section{Preliminary Analysis}\label{sec:analysis}

Leveraging neighboring examples at inference time has been shown to improve the generalization of language models~\cite{knn-lm}, mitigate prompting bias~\cite{knn-prompting}, and improve retrieval-augmented generation~\cite{knn-zero-shot}. Building on these findings, we explore a more focused question: To what extent can semantically similar neighborhood data serve as effective prediction proxies and potentially enhance LLM predictions at test time? 

To understand this, we introduce \textit{neighborhood purity}, which measures how often semantically similar examples share the same label. 
Formally, let $\mathcal{D} = {(x_i, y_i)}_{i=1}^N$ be a set of inputs and corresponding ground truth labels, where $N$ is the total number of data points. We denote the $K$-nearest \textit{neighborhood} of an input $x$ as $\mathcal{N} = \arg\!\operatorname{top}_K \{\mathcal{S}_f\bigl(x, x_i)\,|\; i=0,\dots,N\}$, representing the set of indices corresponding to the most similar instances according to an embedding function $f$. We refer to $x$ as the anchor of the neighborhood and measure the consistency of its neighborhood with \textit{purity} $\phi$, defined as:
\begin{equation}
    \phi\left(\mathcal{N}\right) = \frac{1}{KN} \sum_{i=1}^N\sum_{j \in \mathcal{N}} \mathbf{1}(y_i = y_j)
\end{equation}
Intuitively, purity measures the proportion of instances that share the same label as the anchor.

We conduct our preliminary experiments across eight datasets spanning class granularities from 10 to 150. Detailed dataset descriptions and statistics are provided in Section \ref{sec:experiment_setup} and Table \ref{tab:dataset_statistics}. As shown in Figure~\ref{fig:analysis_neighborhood_purity}, nearest neighbors frequently belong to the same semantic class as the anchor. In the worst case, purity still reaches around 0.3 when $K = 20$ on the GoEmotion dataset. 

\begin{figure}[!t]
    \centering
    \includegraphics[width=\columnwidth]{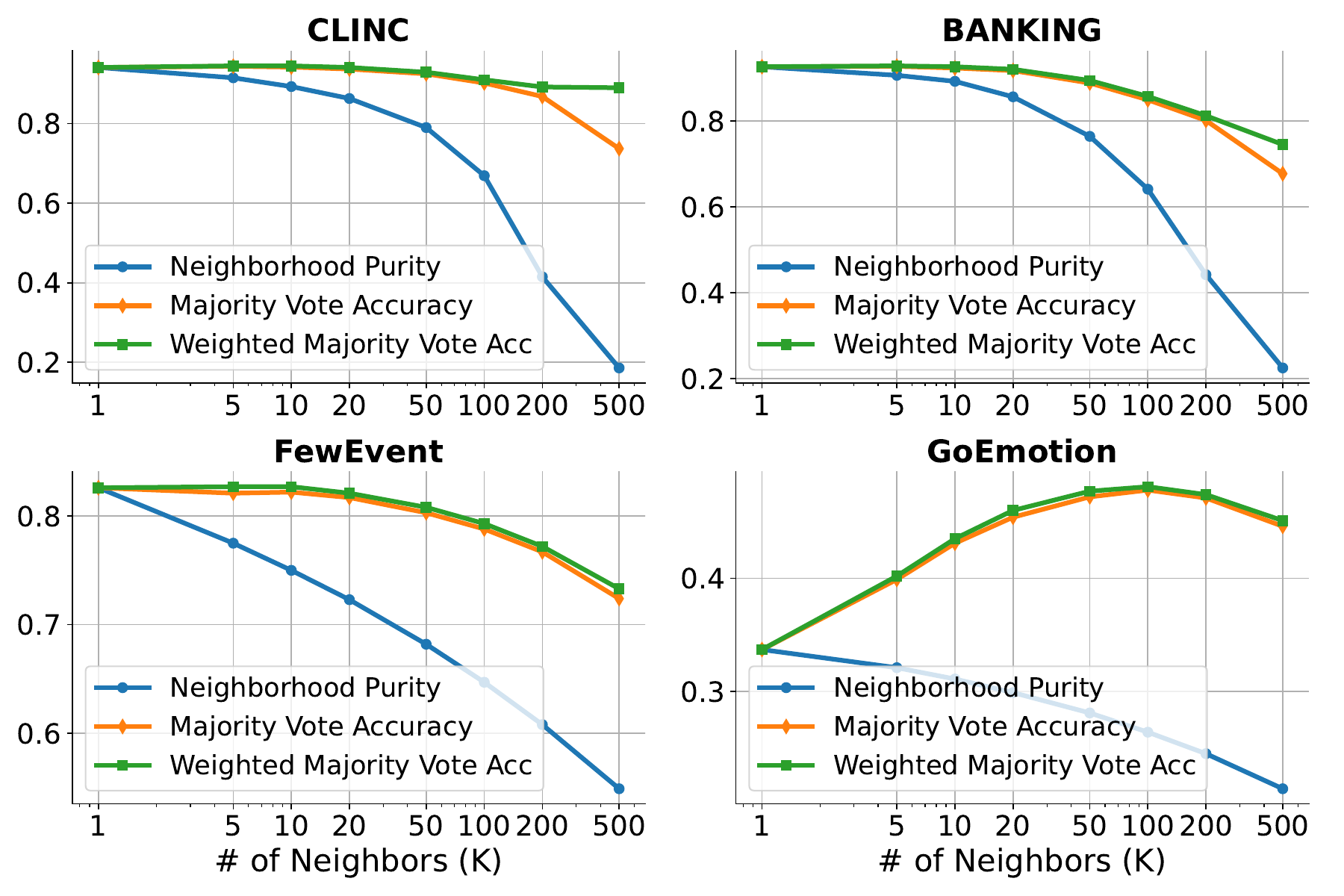} 
    \caption{Majority vote over neighborhood ground-truth labels leads to stable and accurate predictions. Incorporating feature similarity-based weighting further improves stability for large K values by mitigating noise.}
    \label{fig:analysis_majority_vote}
\end{figure}

Then, we ascertain how accurately the aggregation over neighboring ground-truth labels predicts the anchor’s label. To this end, we consider two aggregation strategies: \textit{majority vote} and \textit{weighted majority vote}. Majority vote returns the most frequent class label in the neighborhood: 
\begin{equation} 
    \hat{y}_m(\mathcal{N}) = \argmax_{y} \sum_{i \in \mathcal{N}} \mathbf{1}(y = y_i)
\end{equation}
while weighted majority vote adjusts label counts based on similarity in representation space: 
\begin{equation} 
    \hat{y}_{w}(\mathcal{N}) = \argmax_{y}\sum_{i \in \mathcal{N}} \mathcal{S}_f(x, x_i) \mathbf{1}(y = y_i)
\end{equation}
Figure~\ref{fig:analysis_majority_vote} compares majority-vote accuracy with neighborhood purity across different K values, revealing several key insights: (1) Majority voting over neighboring labels consistently produces accurate anchor predictions; (2) While larger K values decrease neighborhood purity due to noise introduction, majority-vote accuracy remains notably stable, indicating its robustness to the hyperparameter $K$. (3) Similarity-based weighting improves prediction stability for large K values by reducing the impact of less relevant neighbors. 
These findings suggest semantically similar neighborhood data can serve as effective prediction proxies, offering a potential means to enhance LLM predictions at test time.

\section{Method}
Motivated by our findings in Section \ref{sec:analysis}, we propose \model, a test-time computing strategy that leverages neighboring unlabeled data consistency to enhance LLM predictions. Our approach introduces a complementary dimension to test-time computing by integrating signals from unlabeled data during inference. 

\begin{algorithm}[!t]
\caption{\model~algorithm.}
\label{alg:algorithm}
\begin{algorithmic}[1]

\STATE \textbf{Input:} Embedder $f$, test sample $x_0$, unlabeled data $\mathcal{U}=\{u_i\}_{i=1}^{N}$, number of neighbors $K$, threshold $\theta$.

\STATE $z_0 = f(x_0)$, $\mathcal{Z} = \{z_i = f(u_i)\}_{i=1}^N$
\\{\color{gray} \COMMENT{\textit{\small Extract embeddings for test sample and unlabeled data}}}

\STATE $\mathcal{N} = \arg\!\operatorname{top}_K \{\mathcal{S}_f(x_0, x_i)|\; i=0,\dots,N\}$
\\{\color{gray} \COMMENT{\textit{\small Mine top-$K$ neighbors based on similarity, note that test sample $x_0$ is included}}}

\FOR{$k=1$ to $K$}
    \STATE $(y_{\mathcal{N}_k}, \mathit{conf}_{\mathcal{N}_k}) = P_{\mathrm{LLM}}(u_{\mathcal{N}_k})$
    \\{\color{gray} \COMMENT{\textit{\small Prompt LLM to obtain predictions and confidences}}}
    
    \STATE $\color{red}w_k = \operatorname{sim}(z_0, z_{\mathcal{N}_k})$
    \\{\color{gray} \COMMENT{\textit{\small Compute neighbor weights based on similarity}}}
    
    \STATE $\color{blue}c_k = \mathbbm{1}\!(\mathit{conf}_{\mathcal{N}_k} \ge \theta)$ 
    \\{\color{gray} \COMMENT{\textit{\small Filter out unconfident predictions}}}
\ENDFOR

\STATE $y_{\mathrm{final}} \;=\; \underset{y}{\arg\max} \;\sum_{k=1}^{K} {\color{blue}c_k} \, {\color{red}w_k} \,\mathbbm{1}\!(y_{\mathcal{N}_k} = y)$
\\{\color{gray} \COMMENT{\textit{\small Aggregate neighbors' predictions by majority voting}}}

\STATE \textbf{Return} $y_{\mathrm{final}}$

\end{algorithmic}
\end{algorithm}

\subsection{Framework Overview}
TestNUC consists of two key steps:
\begin{itemize}\setlength{\itemsep}{0pt}\setlength{\leftskip}{0pt}\setlength{\parindent}{0pt}
    \item \textbf{Step 1: Neighbor Retrieval.} Identify the top-K nearest neighbors of a test sample based on feature similarity. 
    \item \textbf{Step 2: Collaborative Prediction.} Prompt the LLM to generate predictions for both the test sample and its \( K \) retrieved neighbors. These predictions are combined through a designed aggregation strategy.
\end{itemize}
Note that TestNUC is based on LLM predictions instead of the ground truth label. The intuition behind TestNUC is that samples in close proximity within the embedding space are likely to share similar labels. By incorporating predictions on nearby unlabeled samples, the LLM can better contextualize and refine its decision-making. This approach aims to exploit the consistency of local data structures, effectively using unlabeled examples as an auxiliary signal to boost inference-time performance and reduce the noise and uncertainty associated with isolated predictions.

\subsection{Aggregation Strategy}
The aggregation strategy in Step 2 affects the sensitivity of TestNUC to noise. In this work, we explore three types of aggregation strategies. \\

\noindent \textbf{Naive Majority Voting.} The naive approach simply selects the most consistent answer across the \( K \) unlabeled data predictions. \\

\noindent \textbf{Weighted Majority Voting.} As demonstrated in our analysis in Section \ref{sec:analysis}, when using a large \( K \), neighborhood purity tends to decline rapidly. This indicates that distant neighbors can introduce significant noise and negatively impact the accuracy of majority voting. To mitigate this issue, we additionally use cosine similarity distance between the test sample and its neighbors as weights for majority voting. \\

\noindent \textbf{Filtered Weighted Majority Voting.} The quality of LLM's predictions for neighboring unlabeled data can affect the accuracy of the aggregated results. In this approach, we explore leveraging verbalized confidence to filter out low-quality predictions during majority voting. Specifically, for each unlabeled data, we ask LLM to generate both the prediction and confidence in its predictions and only high confidence predictions are kept for majority voting. \\

\noindent A complete algorithm for Filtered Weighted Majority Voting is presented in Algorithm \ref{alg:algorithm}. The algorithms for the other two voting strategies mentioned above can be obtained by removing the blue- and red-colored code. More complex aggregation strategies can also be explored, such as adding additional distance-based filtering mechanisms or confidence-weighting mechanisms, which we leave for interested researchers to explore.

\begin{table*}[!th]
\centering
\resizebox{\textwidth}{!}{%
\begin{tabular}{@{}llcccccccccc@{}}
\toprule
& & \multicolumn{2}{c}{\textbf{Intent Detection}} & \multicolumn{2}{c}{\textbf{Topic Mining}} & \multicolumn{2}{c}{\textbf{Domain Discovery}} & \multicolumn{1}{c}{\textbf{Type}} & \multicolumn{1}{c}{\textbf{Emotion}} & \\
\cmidrule(lr){3-4} \cmidrule(lr){5-6} \cmidrule(lr){7-8} \cmidrule(lr){9-9} \cmidrule(lr){10-10}  
\textbf{Model} & \textbf{Method} & \textbf{BANKING} & \textbf{CLINC} & \textbf{Reddit} & \textbf{StackEx} & \textbf{MTOP} & \textbf{CLINC(D)} & \textbf{FewEvent} & \textbf{GoEmotion} & \textbf{AVG} \\ \midrule \midrule
GPT-4o-mini & Standard Prompting & 0.652 & 0.792 & 0.534 & 0.482 & 0.896 & 0.536 & 0.630 & 0.378 & 0.613 \\
& Self-Consistency & 0.666 & 0.802 & 0.586 & 0.494 & 0.902 & 0.530 & 0.640 & 0.382 & 0.625 \\
& TestNUC & 0.712 & 0.858 & 0.614 & 0.528 & 0.936 & 0.544 & 0.674 & 0.410 & 0.660 \\
& \cellcolor{gray!18}TestNUC\textdagger & \cellcolor{gray!18}\textbf{0.764} & \cellcolor{gray!18}\textbf{0.864} & \cellcolor{gray!18}\textbf{0.646} & \cellcolor{gray!18}\textbf{0.540} & \cellcolor{gray!18}\textbf{0.948} & \cellcolor{gray!18}\textbf{0.554} & \cellcolor{gray!18}\textbf{0.680} & \cellcolor{gray!18}\textbf{0.414} & \cellcolor{gray!18}\textbf{0.676} \\ \midrule \midrule
Llama-3.1-8B & Standard Prompting & 0.572 & 0.726 & 0.502 & 0.492 & 0.892 & 0.528 & 0.530 & 0.332 & 0.572 \\
& Self-Consistency & 0.620 & 0.774 & 0.564 & 0.526 & 0.902 & 0.518 & 0.564 & 0.340 & 0.601 \\
& TestNUC & 0.694 & 0.806 & 0.618 & 0.558 & 0.934 & 0.528 & 0.596 & 0.356 & 0.636 \\
& \cellcolor{gray!18}TestNUC\textdagger & \cellcolor{gray!18}\textbf{0.724} & \cellcolor{gray!18}\textbf{0.812} & \cellcolor{gray!18}\textbf{0.646} & \cellcolor{gray!18}\textbf{0.576} & \cellcolor{gray!18}\textbf{0.940} & \cellcolor{gray!18}\textbf{0.542} & \cellcolor{gray!18}\textbf{0.614} & \cellcolor{gray!18}\textbf{0.360} & \cellcolor{gray!18}\textbf{0.652} \\ \midrule \midrule
Claude-3-Haiku & Standard Prompting & 0.680 & 0.848 & 0.486 & 0.564 & 0.892 & 0.552 & 0.594 & 0.336 & 0.619 \\
& Self-Consistency & 0.702 & 0.870 & 0.510 & 0.578 & 0.904 & 0.564 & 0.568 & 0.350 & 0.631 \\
& TestNUC & 0.762 & 0.894 & 0.596 & 0.588 & 0.940 & 0.590 & 0.620 & 0.348 & 0.667 \\
& \cellcolor{gray!18}TestNUC\textdagger & \cellcolor{gray!18}\textbf{0.804} & \cellcolor{gray!18}\textbf{0.902} & \cellcolor{gray!18}\textbf{0.612} & \cellcolor{gray!18}\textbf{0.600} & \cellcolor{gray!18}\textbf{0.946} & \cellcolor{gray!18}\textbf{0.622} & \cellcolor{gray!18}\textbf{0.660} & \cellcolor{gray!18}\textbf{0.368} & \cellcolor{gray!18}\textbf{0.689} \\ \midrule \midrule
GPT-4o & Standard Prompting & 0.746 & 0.924 & 0.712 & 0.674 & 0.962 & 0.614 & 0.682 & 0.406 & 0.715 \\
& Self-Consistency & 0.758 & 0.922 & 0.720 & 0.688 & 0.958 & 0.624 & 0.696 & 0.426 & 0.724 \\
&TestNUC & 0.804 & 0.934 & 0.744 & \textbf{0.710} & 0.974 & 0.644 & 0.692 & 0.446 & 0.744 \\
& \cellcolor{gray!18}TestNUC\textdagger & \cellcolor{gray!18}\textbf{0.824} & \cellcolor{gray!18}\textbf{0.940} & \cellcolor{gray!18}\textbf{0.750} & \cellcolor{gray!18}\textbf{0.710} & \cellcolor{gray!18}\textbf{0.978} & \cellcolor{gray!18}\textbf{0.654} & \cellcolor{gray!18}\textbf{0.708} & \cellcolor{gray!18}\textbf{0.464} & \cellcolor{gray!18}\textbf{0.754} \\
\bottomrule
\end{tabular}%
}
\caption{Accuracy comparison with Standard Prompting and Self-Consistency across four diverse LLMs. TestNUC consistently improves the inference performance on all benchmark datasets. $\dagger$ denotes that 50 neighbors are utilized.}
\label{tab:main_compare_sc}
\end{table*}
\begin{table*}[!th]
\centering
\resizebox{\textwidth}{!}{%
\begin{tabular}{@{}llcccccccccc@{}}
\toprule
& & \multicolumn{2}{c}{\textbf{Intent Discovery}} & \multicolumn{2}{c}{\textbf{Topic Mining}} & \multicolumn{2}{c}{\textbf{Domain Discovery}} & \multicolumn{1}{c}{\textbf{Type}} & \textbf{Emotion} & \\
\cmidrule(lr){3-4} \cmidrule(lr){5-6} \cmidrule(lr){7-8} \cmidrule(lr){9-9} \cmidrule(lr){10-10} 
\textbf{Method} & & \textbf{BANKING} & \textbf{CLINC} & \textbf{Reddit} & \textbf{StackEx} & \textbf{MTOP} & \textbf{CLINC(D)} & \textbf{FewEvent} & \textbf{GoEmotion} & \textbf{AVG} \\ \midrule \midrule
\textbf{KNN-ICL} & & 0.664 & 0.768 & 0.670 & 0.520 & 0.942 & 0.518 & 0.570 & 0.386 & 0.630 \\
w. TestNUC & & 0.762 & 0.832 & 0.728 & 0.566 & 0.960 & 0.544 & 0.606 & 0.410 & 0.676 \\
Improvement & & {\color[HTML]{34a854}\textbf{ 14.76\%}} & {\color[HTML]{34a854}\textbf{ 8.33\%}} & {\color[HTML]{34a854}\textbf{ 8.66\%}} & {\color[HTML]{34a854}\textbf{ 8.85\%}} & {\color[HTML]{34a854}\textbf{ 1.91\%}} & {\color[HTML]{34a854}\textbf{ 5.02\%}} & {\color[HTML]{34a854}\textbf{ 6.32\%}} & {\color[HTML]{34a854}\textbf{ 6.22\%}} & {\color[HTML]{34a854}\textbf{ 7.51\%}} \\ \midrule \midrule
\textbf{KNN-ICL-P} & & 0.702 & 0.870 & 0.620 & 0.556 & 0.922 & 0.548 & 0.624 & 0.416 & 0.657 \\
w. TestNUC & & 0.768 & 0.894 & 0.672 & 0.584 & 0.960 & 0.584 & 0.654 & 0.444 & 0.695 \\
Improvement & & {\color[HTML]{34a854}\textbf{ 9.40\%}} & {\color[HTML]{34a854}\textbf{ 2.76\%}} & {\color[HTML]{34a854}\textbf{ 8.39\%}} & {\color[HTML]{34a854}\textbf{ 5.04\%}} & {\color[HTML]{34a854}\textbf{ 4.12\%}} & {\color[HTML]{34a854}\textbf{ 6.57\%}} & {\color[HTML]{34a854}\textbf{ 4.81\%}} & {\color[HTML]{34a854}\textbf{ 6.73\%}} & {\color[HTML]{34a854}\textbf{ 5.98\%}} \\ \midrule \midrule
\textbf{Self-Consistency} & & 0.666 & 0.802 & 0.586 & 0.494 & 0.902 & 0.530 & 0.640 & 0.382 & 0.625 \\
w. TestNUC & & 0.750 & 0.878 & 0.706 & 0.562 & 0.928 & 0.566 & 0.670 & 0.420 & 0.685 \\
Improvement & & {\color[HTML]{34a854}\textbf{ 12.61\%}} & {\color[HTML]{34a854}\textbf{ 9.48\%}} & {\color[HTML]{34a854}\textbf{ 20.48\%}} & {\color[HTML]{34a854}\textbf{ 13.77\%}} & {\color[HTML]{34a854}\textbf{ 2.88\%}} & {\color[HTML]{34a854}\textbf{ 6.79\%}} & {\color[HTML]{34a854}\textbf{ 3.69\%}} & {\color[HTML]{34a854}\textbf{ 9.95\%}} & {\color[HTML]{34a854}\textbf{ 9.56\%}} \\ \midrule \midrule
\textbf{Best-of-N} & & 0.662 & 0.814 & 0.606 & 0.492 & 0.902 & 0.544 & 0.620 & 0.378 & 0.627 \\
w. TestNUC & & 0.758 & 0.880 & 0.706 & 0.568 & 0.954 & 0.564 & 0.696 & 0.412 & 0.692 \\
Improvement & & {\color[HTML]{34a854}\textbf{ 14.50\%}} & {\color[HTML]{34a854}\textbf{ 8.11\%}} & {\color[HTML]{34a854}\textbf{ 16.50\%}} & {\color[HTML]{34a854}\textbf{ 15.45\%}} & {\color[HTML]{34a854}\textbf{ 5.76\%}} & {\color[HTML]{34a854}\textbf{ 3.68\%}} & {\color[HTML]{34a854}\textbf{ 12.26\%}} & {\color[HTML]{34a854}\textbf{ 8.99\%}} & {\color[HTML]{34a854}\textbf{ 10.36\%}} \\ \midrule \midrule
\textbf{Weighted Best-of-N} & & 0.658 & 0.820 & 0.602 & 0.484 & 0.900 & 0.532 & 0.612 & 0.372 & 0.623 \\
w. TestNUC & & 0.752 & 0.876 & 0.710 & 0.558 & 0.938 & 0.566 & 0.672 & 0.422 & 0.687 \\
Improvement & & {\color[HTML]{34a854}\textbf{ 14.29\%}} & {\color[HTML]{34a854}\textbf{ 6.83\%}} & {\color[HTML]{34a854}\textbf{ \textbf{17.94\%}}} & {\color[HTML]{34a854}\textbf{ 15.29\%}} & {\color[HTML]{34a854}\textbf{ 4.22\%}} & {\color[HTML]{34a854}\textbf{ 6.39\%}} & {\color[HTML]{34a854}\textbf{ 9.80\%}} & {\color[HTML]{34a854}\textbf{ 13.44\%}} & {\color[HTML]{34a854}\textbf{ 10.32\%}} \\
\bottomrule
\end{tabular}%
}
\caption{TestNUC can significantly enhance various existing test-time computing approaches - both those that prepend demonstrations at the input level (ICL-based) and those that do sampling and “post-hoc” candidate refinements at the output level (Self-Consistency, Best-of-N). The relative improvement is visualized in Figure \ref{fig:radar_plot}.}
\label{tab:main_enhance}
\end{table*}



\begin{figure*}[!th]
    \centering
    \includegraphics[width=1\textwidth]{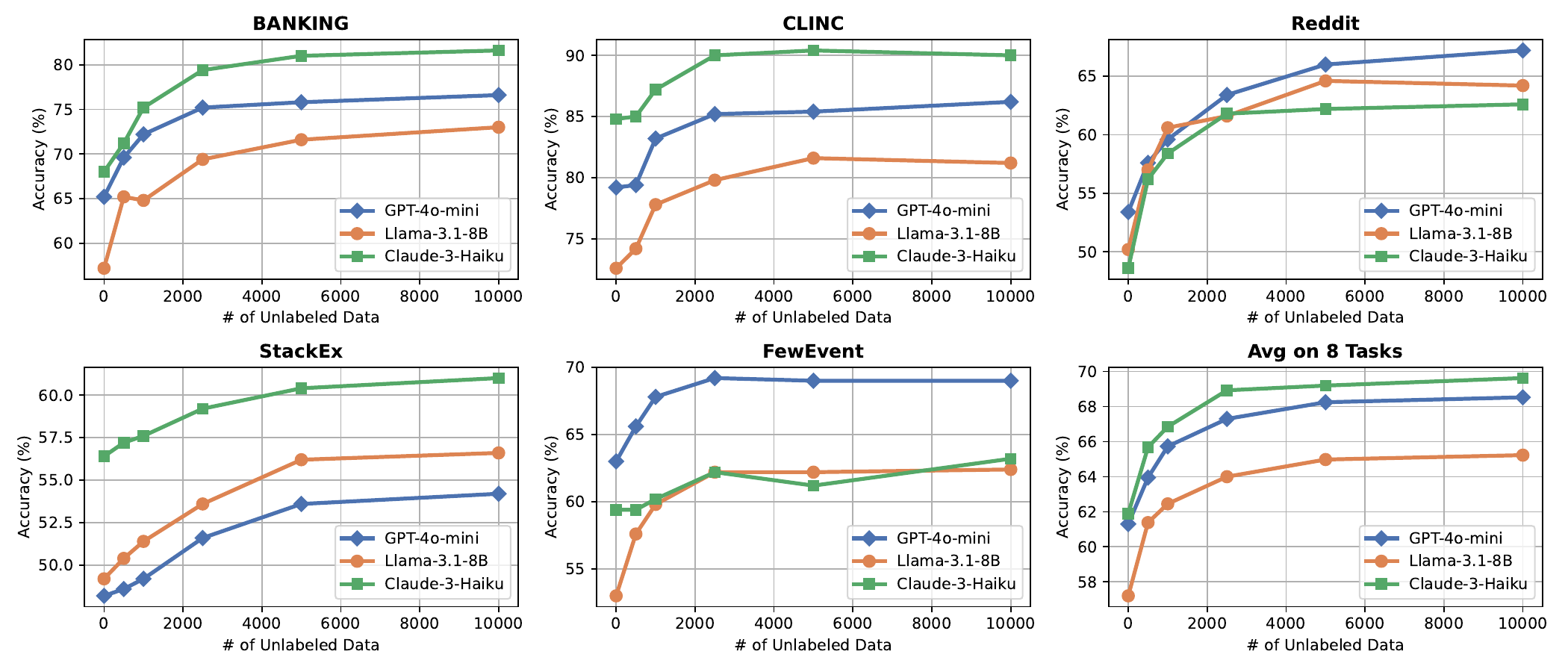} 
    \caption{Increasing the amount of unlabeled data consistently boosts performance across all evaluated LLMs and datasets. The scaling trends are more distinctly visible in the logarithmic version of the figure (Figure \ref{fig:influence_num_unlabeled_data_log}).}
    \label{fig:influence_num_unlabeled_data}
\end{figure*}

\begin{figure*}[!th]
    \centering
    \includegraphics[width=1\textwidth]{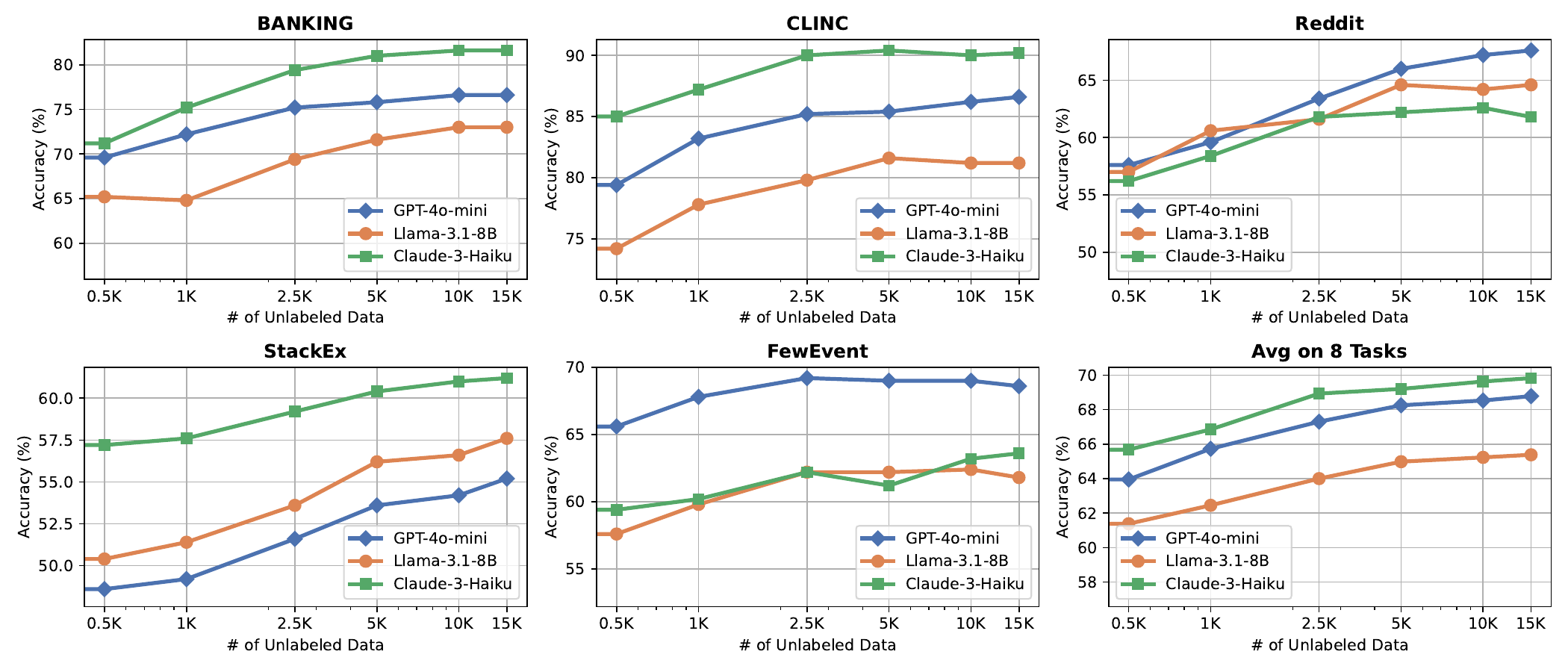} 
    \caption{The logarithmic version of Figure \ref{fig:influence_num_unlabeled_data}.}
    \label{fig:influence_num_unlabeled_data_log}
\end{figure*}

\section{Experiments}
\label{sec:experiment}
\subsection{Experiment Setup}
\label{sec:experiment_setup}

\noindent \textbf{Tasks and Datasets.} We consider eight datasets across diverse tasks with various perspectives and granularities as follows.
\begin{itemize}[leftmargin=*]
\setlength{\itemsep}{0pt}\setlength{\leftskip}{0pt}\setlength{\parindent}{0pt}
    \item \textbf{Intent Detection.} Intent detection aims to discover fine-grained intents in customer utterances. We use BANKING \cite{banking} and CLINC \cite{clinic} for evaluation.

    \item  \textbf{Topic Mining.} We use Reddit and StackExchange from MTEB \cite{muennighoff-etal-2023-mteb} and ClusterLLM \cite{zhang-etal-2023-clusterllm} to evaluate models' ability to categorize discussion topics.

    \item  \textbf{Domain Discovery.} For this task, we use MTOP \cite{mtop} and CLINC(D) \cite{zhang-etal-2023-clusterllm} to allow evaluations of models' capability in discovering domain-specific knowledge.

    \item  \textbf{Type Discovery.} We use the FewEvent dataset \cite{FewEvents} that focuses on extracting event types from the given text and event triggers. 

    \item  \textbf{Emotion Recognition.} We use GoEmotion \cite{goemotions}, which is a dataset of Reddit comments labeled with fine-grained emotions, such as amusement, fear and gratitude.
\end{itemize}
\noindent Dataset statistics are summarized in Appendix \ref{appendix:dataset}.
\\

\noindent \textbf{Baselines. } We consider three types of baselines: 

\noindent \textbf{\ding{182}} \textbf{Standard Prompting}, which prompts the LLM in a standard way to select a label from the provided options to a test sample. The details of the prompt template are available in Appendix \ref{sec:prompt_template}. \\

\noindent \textbf{\ding{183}} Test-time computing approaches that operate \textit{at the input level} by augmenting the given prompt with additional demonstrations to enhance inference performance. 
Since our proposed method combines decisions based on similar examples, we compare it with two varieties of in-context learning counterparts: \textbf{TopK-ICL} \cite{peng-etal-2024-revisiting}, where the input text of the nearest neighbors of the test example are added to the prompt as context information. \textbf{TopK-ICL-P}, where we additionally append each neighbor's Standard Prompting prediction result to its text as demonstrations. \\

\noindent \textbf{\ding{184}} Test-time computing approaches that operate \textit{at the output level} through multiple candidate answer sampling and aggregation to boost output quality. For this category, we consider three representative approaches: \textbf{Self-Consistency} \cite{self-consistency}, \textbf{Best-of-N} \cite{test-time-compute, beeching2024scalingtesttimecompute}, and \textbf{Weighted Best-of-N} \cite{beeching2024scalingtesttimecompute}. 
Specifically, Best-of-N selects the most confident predictions out of multiple predictions based on the LLM's own verbalized confidence \cite{verbalized-confidence}. Weighted Best-of-N aggregates the decisions by assigning weights based on their respective confidence score.\\

\noindent \textbf{Implementation Details. } We utilize both open-sourced and close-sourced LLMs with varying scales: GPT-4o-mini, GPT-4o \cite{gpt4o}, Llama-3.1-8B \cite{dubey2024llama}, Claude 3 Haiku \cite{claude3}. We set temperature $T=0.7$  and Top-p $= 1.0$ for sampling decoding for all evaluated language models. By default, the number of candidate answers $N$  we sampled for Self-Consistency, Best-of-N and Weighted Best-of-N is 10. Similarly, the number of retrieved neighbors, i.e., $K$, for TopK-ICL, TopK-ICL-P, and our TestNUC is 10 unless stated otherwise. We adopt NV-Embed-v2-7B \cite{lee2024nv} as the embedding model for all methods. Due to resource constraints, we randomly sample 500 data points from each dataset for evaluation and use the remaining for neighboring sample retrieval.

\subsection{Main Results}


\noindent \textbf{Comparison with Standard Prompting and Self-Consistency. } Table \ref{tab:main_compare_sc} presents the comparison results with Standard Prompting and Self-Consistency across four large language models. It can be observed that \model~significantly improves the inference performance of four large language models on all eight evaluated datasets over standard prompting. TestNUC can also outperform Self-Consistency when utilizing the same amount of sampling paths and neighboring unlabeled data (i.e., $K$=10 in both cases). For example, TestNUC surpasses Self-Consistency by 5.87\% on average when using Llama-3.1-8B model and 5.48\% on average when using GPT-4o-mini. Besides, \model performance can be further boosted by utilizing more neighboring unlabeled data. TestNUC$\dagger$, which utilizes 50 neighbors, can improve the performance over standard prompting up to 11.35\% on average across 8 datasets when using Claude-3-Haiku. Additionally, performance improvements are observed across all four language models, even in the already powerful GPT-4o model. 


\begin{table*}[!th]
\centering
\resizebox{\textwidth}{!}{%
\begin{tabular}{@{}lcccccccccc@{}}
\toprule
\textbf{Aggregation Strategy} & \textbf{BANKING} & \textbf{CLINC} & \textbf{Reddit} & \textbf{StackEx} & \textbf{MTOP} & \textbf{CLINC(D)} & \textbf{FewEvent} & \textbf{GoEmotion} & \textbf{AVG} \\ \midrule \midrule
Standard Prompting & 0.652 & 0.792 & 0.534 & 0.482 & 0.896 & 0.536 & 0.630 & 0.378 & 0.613 \\ \midrule \midrule
Naive Majority Voting & 0.756 & 0.862 & 0.644 & 0.538 & 0.948 & 0.550 & 0.668 & 0.392 & 0.670 \\
Weighted Majority Voting (Distance) & 0.764 & 0.864 & 0.646 & 0.540 & 0.948 & 0.554 & 0.680 & 0.414 & 0.676 \\
Weighted Majority Voting (Distance \& Confidence) & \textbf{0.768} & 0.870 & 0.656 & 0.542 & 0.948 & 0.552 & 0.684 & \textbf{0.416} & 0.680 \\
Filtered Weighted Majority Voting & 0.762 & \textbf{0.876} & \textbf{0.688} & \textbf{0.542} & \textbf{0.954} & \textbf{0.572} & \textbf{0.688} & 0.410 & \textbf{0.687} \\
\bottomrule
\end{tabular}%
}
\caption{Comparison of aggregation strategies across diverse datasets. Naive majority voting already significantly improves accuracy over standard prompting. Weighted Majority Voting with distance and confidence further enhances performance, and filtering low-confidence predictions achieves the highest average result.}
\label{tab:voting_comparison}
\end{table*}

\noindent \textbf{TestNUC can enhance various existing test-time computing approaches. } The results are shown in Table \ref{tab:main_enhance}.  Across all baselines and all datasets, incorporating TestNUC boosts performance. The average improvements range from about +6\% (TopK-ICL-P) to +10\% (Best-of-N and Weighted Best-of-N), indicating that TestNUC is complementary to diverse inference strategies—both those that prepend demonstrations at the input level (ICL-based) and those that do sampling and “post-hoc” candidate refinements at the output level (Self-Consistency, Best-of-N). Methods that work at the output level (e.g., Self-Consistency, Best-of-N) present larger average gains (9–10\%), compared to input-level approaches (6–7\%). 
The biggest performance gains often appear on the Topic Mining tasks (e.g., Reddit, StackEx), where improvements can exceed +15–20\%.
This suggests that adding \model~is especially helpful in scenarios involving more open‐ended or noisy textual inputs, where post-hoc aggregations can more effectively disambiguate the model’s initial outputs.


\begin{table}[!t]
\centering
\resizebox{\columnwidth}{!}{
\begin{tabular}{@{}lccccc@{}}
\toprule
\textbf{Embedder} & \textbf{BANKING} & \textbf{CLINC} & \textbf{Reddit} & \textbf{MTOP} & \textbf{AVG} \\ \midrule \midrule
Standard Prompting & 0.652 & 0.792 & 0.534 & 0.896 & 0.719 \\ \midrule \midrule
all-MiniLM-L12-v2-120M & 0.706 & 0.832 & 0.584 & 0.928 & 0.763 \\
all-distilroberta-v1-290M & 0.690 & 0.840 & 0.586 & 0.938 & 0.764 \\
all-mpnet-base-v2-420M & 0.712 & 0.852 & 0.586 & 0.942 & 0.773 \\ \midrule \midrule
gte-Qwen2-1.5B-instruct & 0.694 & 0.844 & 0.614 & 0.946 & 0.775 \\
stella-en-400M-v5 & 0.728 & 0.834 & 0.618 & 0.946 & 0.782 \\
NV-Embed-v2-7B & \textbf{0.764} & \textbf{0.864} & \textbf{0.646} & \textbf{0.948} & \textbf{0.806} \\
\bottomrule
\end{tabular}}
\caption{Results with varying embedding models. TestNUC is effective on diverse embedding models from different companies and of different sizes.}
\label{tab:embedder_comparison}
\end{table}

\vspace{10pt}
\section{Additional Studies}

\subsection{Influence of Unlabeled Data Size}

\noindent \textbf{Increasing unlabeled data helps boost performance across tasks.}
Figure~\ref{fig:influence_num_unlabeled_data} reports the linear-scale results on BANKING, CLINC, Reddit, StackEx, FewEvent, and an overall average across eight tasks. In all cases, increasing the unlabeled set yields notable accuracy improvements for GPT-4o-mini, Llama-3.1-8B, and Claude-3-Haiku. Adding even a modest number (e.g., 500–1k) of unlabeled instances yields substantial accuracy gains, especially on BANKING and Reddit.
However, improvements taper off after roughly 8–10k unlabeled samples, suggesting a saturation point where additional unlabeled data provides diminishing returns. 
The log-scale plots in Figure~\ref{fig:influence_num_unlabeled_data_log} further highlight these trends, confirming that the utility of unlabeled examples gradually diminishes but still delivers meaningful improvements up to the 10K–15K range. This pattern holds consistently across all tasks, confirming that increasing unlabeled data universally improves performance but with predictable saturation effects. \\

\vspace{-5pt}
\subsection{Aggregation Strategy Comparison}

We find that \textbf{naive majority voting greatly surpasses standard prompting performance, with advanced strategies further enhancing results.} As shown in Table \ref{tab:voting_comparison}, simply aggregating multiple predictions with naive majority voting can already boost average accuracy significantly from 0.613 to 0.670. Introducing distance and confidence weighting further refines these gains from the average to 0.680. Finally, filtering out low-confidence predictions yields the highest performance, although on certain tasks (e.g., GoEmotion), Weighted Majority Voting (Distance \& Confidence) can be more effective, suggesting that a carefully tuned confidence threshold may be necessary for each dataset. \\

\vspace{-5pt}
\subsection{Varying Embedding Models}
\noindent \textbf{TestNUC works on different sizes of embedders. } The embedders used to generate data embeddings for neighbor retrieval play a crucial role in the success of TestNUC. In this work, we explore diverse embedding models, including public encoders from different companies and embedders of different sizes ranging from 120M to 7B. 
As shown in Table \ref{tab:embedder_comparison}, TestNUC is effective when applied to various embedding models. Not surprisingly, TestNUC achieves significant improvements with larger and more advanced embedders, such as NV-Embed-v2-7B, which records the highest average performance (0.755) and excels across all datasets. 
Mid-sized embedders like stella-en-400M-v5 (0.738) and gte-Qwen2-1.5B-instruct (0.732) also perform well, demonstrating that TestNUC can effectively leverage diverse embedding architectures. Even smaller models, such as all-MiniLM-L12-v2-120M (0.720), deliver competitive results over standard prompting (0.682), showcasing TestNUC's robustness across varying model sizes and complexities.

\begin{figure*}[!th]
    \centering
    \includegraphics[width=1\textwidth]{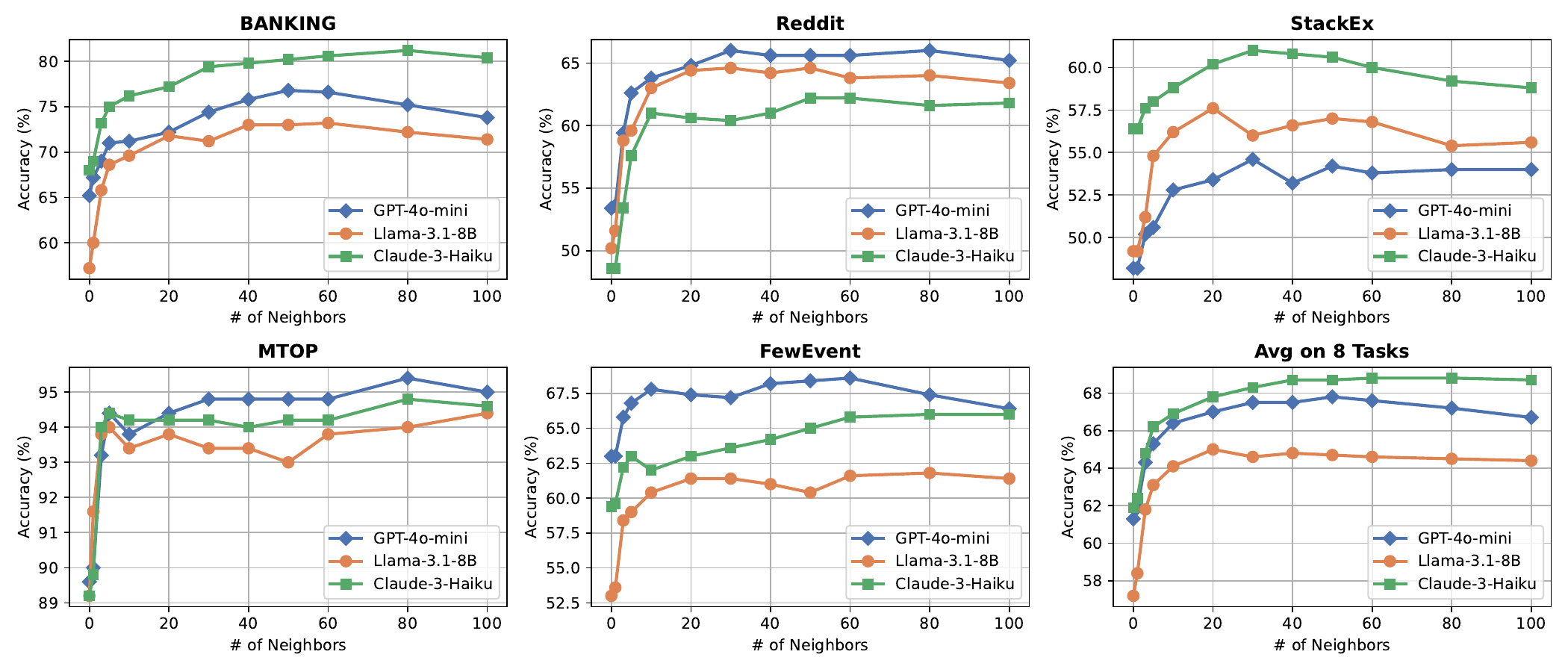} 
    \caption{Influence of the number of neighbors. The results show that even a small set of neighbors can significantly boost performance for all three LLMs, significantly surpassing their zero-neighbor baselines.}
    \label{fig:ablation_num_neighbors}
\end{figure*}

\subsection{Influence of Neighbor Size}
\noindent \textbf{Influence of the Number of Neighboring Unlabeled Data. } Figure~\ref{fig:ablation_num_neighbors} shows the results of TestNUC from GPT-4o-mini, Llama-3.1-8B, and Claude-3-Haiku as the number of neighbors increases. 
The results show that even a small set of neighbors can significantly boost performance for all three models, significantly surpassing their zero-neighbor baselines. Additionally, all three models generally benefit from increasing the number of neighbors from 0 to 60, although the gains tend to plateau after approximately 40–60 neighbors. Notably, the performance of GPT-4o-mini and Llama-3.1-8B slightly decreases when the number of neighbors increases from 60 to 80 on certain datasets, likely due to the introduction of more noisy neighbors. In contrast, Claude-3-Haiku often achieves higher accuracies with relatively larger neighborhood sizes (e.g., 60–100), indicating greater robustness to noise.

\section{Related Work}

\subsection{Test-Time Computing}
Test-time compute \cite{test-time-compute} improves LLM performance by modifying the prediction distribution during test time. Such modification is usually accompanied with extra computational cost. Instead of decoding greedily, the model may sample multiple decoding paths before aggregating them into a response. Chain of Thought \cite{chain-of-thought} modifies the output distribution through hand-crafted prompts that contain reasoning chains. 
Self-consistency \cite{self-consistency} samples multiple chain-of-thought paths and aggregate the sample with majority voting. \cite{ensemble-of-prompt} observed improved accuracy and robustness by querying the model with semantically equivalent prompts before responding with the majority answer. 
\cite{TopK} uses sentence embeddings to retrieve k-nearest-neighbor demonstration for in-context learning. \cite{DPP} retrieves relevant and diverse demonstrations by training a model that predicts the relevance of a demonstration via contrastive learning \cite{SimCLR}. 
Our work is directly inspired by the KNN method proposed by \cite{TopK}. Later work has revealed that similarity based demonstration retrieval improves in-context learning because LLMs attend to the most similar demonstration during few-shot prompting \cite{anchor-words}. Instead of using similar demonstrations for in-context learning, we explore using them as near neighbors in the fashion of non-parametric prediction.

\subsection{In-Context Learning}
Apart from Chain-of-Though, many work explore the possibility of using self-generated content by the LLM to aid with reasoning or classification. STaR \cite{STaR} iteratively add self-generated rationales that are proved correct by a verifier to the exist pool of demonstrations. A significant limitation of STaR is that it relies on knowing the correct answer to the questions the LLM is generating rationale for. Our method simply make predictions for neighboring examples, which does not require ground truth labels. Auto-CoT\cite{Auto-CoT} uses self-generated rationales as demonstrations for similar inputs. The generated data by Auto-CoT incurs a quadratically scaling overhead to the final prediction. Our proposed method only incurs a linearly scaling overhead due to the nature of nearest-neighbor algorithm. Self-ICL \cite{self-icl} generated its own demonstration and their pseudo-labels and uses them as demonstrations. We disagree with Self-ICL's premise that even unlabeled data are hard to come by in realistic settings, and posit that unlabeled data are abundant and inexpensive to obtain \cite{zou-etal-2023-decrisismb, zou2023semi}. Thus, self-generated demonstration inputs are unnecessary. Like Auto-CoT, Self-ICL's test-time compute overhead also scales quadratically. Lastly, Auto-CoT, STaR, and Self-ICL all focuses on reasoning tasks, whereas our work primarily focuses on classification tasks.

\section{Conclusion}
In this work, we introduced TestNUC, a simple yet effective approach that leverages the consistency of neighboring unlabeled data to enhance test-time predictions in large language models. 
Extensive experiments across eight datasets and multiple LLMs demonstrate that TestNUC consistently outperforms baselines like standard prompting and self-consistency. It can be seamlessly integrated with existing methods such as TopK-ICL, self-consistency, and best-of-N to yield further gains. These results highlight the practical value of leveraging unlabeled data during inference, which not only boosts label consistency but also offers a scalable path to better generalization in real-world applications where labeled data may be scarce. 





\section{Limitation}
Our evaluation of TestNUC is limited to classification tasks and does not include generative tasks. We leave this extension for future work. Due to computational resource constraints and limited budgets, we did not evaluate recent powerful reasoning models such as o3-mini and DeepSeek-R1.

\section*{Acknowledgements}

This work is supported in part by NSF under grants III-2106758, and POSE-2346158. 

\bibliography{custom}

\clearpage
\appendix

\section{Dataset Statistics}
\label{appendix:dataset}
Table \ref{tab:dataset_statistics} provides the dataset statistics summary for all evaluated datasets.

\begin{table}[!ht]
\centering
\resizebox{\columnwidth}{!}{%
\begin{tabular}{@{}ccccc@{}}
\toprule
\textbf{Task} & \textbf{Dataset} & \textbf{\# Classes} & \textbf{Total} & \textbf{Test} \\ \midrule \midrule
\multirow{2}{*}{Intent Detection} & BANKING & 77 & 10,003 & 500 \\
 & CLINC & 150 & 15,000 & 500 \\ \midrule \midrule
\multirow{2}{*}{Topic Mining} & Reddit & 50 & 25,000 & 500 \\
 & StackExchange & 121 & 25,000 & 500 \\ \midrule \midrule
\multirow{2}{*}{Domain Discovery} & MTOP & 11 & 15,667 & 500 \\
 & CLINC(D) & 10 & 15,000 & 500 \\ \midrule \midrule
Type Discovery & FewEvent & 34 & 18909 & 500 \\ \midrule
\midrule
Emotion Detection & GoEmotion & 27 & 23485 & 500 \\ \bottomrule
\end{tabular}%
}
\caption{Dataset statistics.}
\label{tab:dataset_statistics}
\end{table}

\begin{figure*}[!t]
    \centering
    \includegraphics[width=\textwidth]{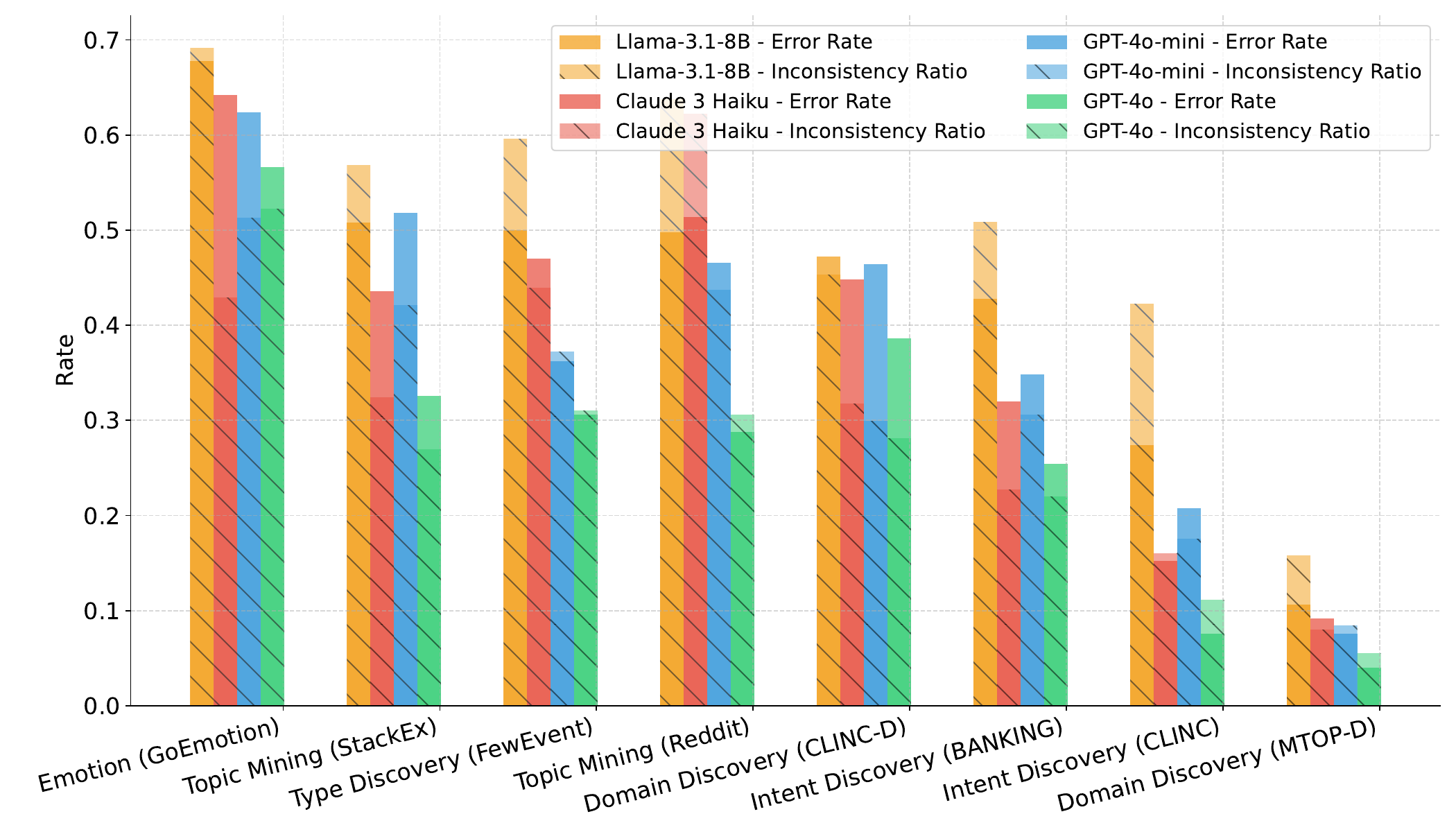} 
    \caption{LLM predictions can be inaccurate and unstable.} \label{fig:analysis_llm_predictions}
\end{figure*}


\section{Prompt Template}
\label{sec:prompt_template}
The prompt template we used in the experiments is listed below. Note that we use the same prompt template for all methods for fair comparisons.

{ \begin{tcolorbox}[
    colback=gray!4,     
    colframe=black!55,
    title=Prompt Template
]
Instruction: Please select a label from the provided options for the following testing samples and also show your confidence in the label assignment by providing a probability between 0 and 1. \\

Label Options: \texttt{[A List of Labels]}. \\

== Testing Samples == \\
\texttt{[Testing Samples]}
\end{tcolorbox}
}

\section{LLM Predictions Can Be Inaccurate and Unstable}
Figure~\ref{fig:analysis_llm_predictions} demonstrates the error rate and inconsistency ratio of predictions by different LLMs on diverse datasets. The inconsistency ratio here refers to the proportion of prediction changes when an LLM is rerun \(N\) times for the same input query across the entire dataset. The results are obtained using standard zero-shot prompting with a temperature of 0.7 and \(N = 10\). It can be observed that even in standard text classification tasks, LLMs can produce inaccurate and inconsistent prediction results for ambiguous or challenging data points.

\section{Robustness in Adversarial Scenarios}

In real-world applications, suitable in-distribution data may not always be available, and retrieved samples could be out-of-distribution. This section demonstrates the robustness of our TestNUC method in such adversarial scenarios. We conducted an adversarial experiment by replacing in-distribution samples in the \textsc{Banking} dataset with out-of-distribution samples (i.e., outliers) from the \textsc{Reddit} dataset. To create sufficiently challenging scenarios, we replaced 60\% and 75\% of the in-distribution samples with OOD samples. As shown in the Table \ref{tab:ood}, even with 60\%--75\% OOD samples present, TestNUC still significantly improves baseline performance across many cases in these highly noisy scenarios, demonstrating its robustness and effectiveness.

Furthermore, when using Weighted Majority Voting (WMV) instead of Naive Majority Voting (NMV), TestNUC's performance and robustness can be further enhanced. This is because OOD samples are likely to have lower semantic similarity with the test sample compared to in-distribution samples, and thus the model can assign lower weights to OOD samples when using WMV. This is also consistent with our previous findings that WMV outperforms NMV in most cases.

\begin{table}[!h]
\resizebox{\columnwidth}{!}{%
\begin{tabular}{@{}ccccc@{}}
\toprule
OOD Ratios & \multicolumn{2}{c}{60\%} & \multicolumn{2}{c}{75\%} \\ \midrule
\# Neighbors & NMV & WMV & NMV & WMV \\ \midrule
0 & 63.7±0.6 & 63.7±0.6 & 63.7±0.6 & 63.7±0.6 \\
\midrule
3 & 67.9±0.7 & 68.2±0.9 & 68.3±0.7 & 69.2±0.9 \\
5 & 69.6±0.8 & 70.6±0.8 & 70.3±0.7 & 71.1±0.6 \\
10 & 72.3±0.8 & 72.8±0.6 & 73.7±0.7 & 74.0±0.5 \\
20 & 74.7±0.3 & 75.0±0.4 & 73.3±0.7 & 73.3±0.8 \\
30 & 75.9±0.7 & 75.6±0.6 & 72.1±0.6 & 72.8±0.7 \\
40 & 75.7±0.8 & 76.3±0.6 & 69.1±0.8 & 70.8±0.9 \\
50 & 74.7±0.6 & 75.2±0.6 & 67.0±0.7 & 69.6±0.8 \\
60 & 73.5±0.6 & 73.7±0.5 & 62.0±0.8 & 66.0±0.7 \\
80 & 71.3±0.6 & 72.1±0.7 & 54.9±0.6 & 61.7±0.8 \\
100 & 67.3±0.7 & 70.1±0.7 & 48.8±0.6 & 59.0±0.7 \\ \bottomrule
\end{tabular}%
}
\caption{
Performance of TestNUC under adversarial conditions with out-of-distribution (OOD) samples.
}
\label{tab:ood}
\end{table}

\section{Runtime and Cost Analysis}

This section presents an estimated analysis of the trade-off between gains in accuracy (average on all 8 tasks) and the compute/runtime cost required for the approach in different settings. The following table summarizes the estimated runtime and cost per sample using GPT-4o-mini. The number of retrieved neighbors and sampled candidate answers is set to 10 by default, unless otherwise specified.

We include two variants of TestNUC: TestNUC and TestNUC-S. TestNUC-S is an efficient implementation of TestNUC that pre-computes and stores the embeddings and predictions of previously queried or seen samples. Thus, when a new sample arrives, TestNUC-S can simply retrieve the embeddings and predictions of the nearest neighbors from the stored set and use them to generate the final label without additional LLM calls—where the runtime cost for retrieval is negligible compared to querying the LLM. As shown in Table \ref{tab:runtime_cost_analysis}, when increasing the number of retrieved neighbors (K=5 to K=50), TestNUC can greatly improves performance although at the cost of increased runtime and cost. When using the same number of retrieved neighbors (K=10), TestNUC is more efficient than the best-performing baseline KNN-ICL-P, which incurs a computational cost that scales quadratically with K, whereas TestNUC incurs only a linear cost. Moreover, the efficient implementation TestNUC-S can significantly reduce runtime cost and is also a very practical solution for real-world applications, as storing the embeddings and queries of previously queried samples is both quite common and cheap in practice.

\begin{table}[!th]
\resizebox{\columnwidth}{!}{%
\begin{tabular}{@{}lccc@{}}
\toprule
 & \textbf{Runtime (s)} & \textbf{Cost (\$)} & \textbf{Performance} \\ \midrule
Standard Prompting & 0.6820 & \$0.000028 & 0.613 \\
w. TestNUC (K=5) & 3.4100 & \$0.000140 & 0.648 \\
w. TestNUC (K=10) & 6.8200 & \$0.000280 & 0.660 \\
w. TestNUC (K=50) & 34.1000 & \$0.001400 & 0.676 \\
\textbf{w. TestNUC-S (K=5)} & \textbf{0.6842} & \textbf{\$0.000028} & \textbf{0.648} \\
\textbf{w. TestNUC-S (K=10)} & \textbf{0.6843} & \textbf{\$0.000028} & \textbf{0.660} \\
\textbf{w. TestNUC-S (K=50)} & \textbf{0.6847} & \textbf{\$0.000028} & \textbf{0.676} \\ \midrule
KNN-ICL & 0.6972 & \$0.000085 & 0.630 \\
KNN-ICL-P & 7.5510 & \$0.000371 & 0.657 \\
Self-Consistency & 6.8200 & \$0.000280 & 0.625 \\
Best-of-N & 6.8200 & \$0.000280 & 0.627 \\ \bottomrule
\end{tabular}%
}
\caption{Runtime and cost analysis. TestNUC improves performance with more neighbors at higher cost, while TestNUC-S achieves the same gains with negligible runtime by reusing stored embeddings and predictions.}
\label{tab:runtime_cost_analysis}
\end{table}

\end{document}